\documentclass[letterpaper]{article} %
\usepackage{aaai23}  %
\usepackage{times}  %
\usepackage{helvet}  %
\usepackage{courier}  %
\usepackage[hyphens]{url}  %
\usepackage{graphicx} %
\usepackage{natbib}  %
\usepackage{caption} %
\usepackage{algorithm}
\usepackage{algorithmic}

\usepackage{multirow}
\usepackage{amsmath} 
\usepackage{amssymb}
\usepackage{mathtools}
\usepackage{amsthm}
\usepackage{pifont} %
\newcommand{\xmark}{\ding{55}}
\newcommand{\cmark}{\ding{51}}
\theoremstyle{plain}
\newtheorem{theorem}{Theorem}

\newtheorem{lemma}[theorem]{Lemma}

\theoremstyle{definition}
\newtheorem{definition}[theorem]{Definition}

\theoremstyle{remark}

\usepackage{xspace} %

\newcommand{\eat}[1]{}

\newcommand{\ie}[0]{\emph{i.e.},\xspace}

\newcommand{\x}[0]{\ensuremath{\mathbf{x}}\xspace}
\newcommand{\y}[0]{\ensuremath{\mathbf{y}}\xspace}
\newcommand{\param}[0]{\ensuremath{\boldsymbol{\theta}}\xspace}
\usepackage[capitalize,noabbrev]{cleveref}
\usepackage{microtype}
\usepackage{graphicx}
\usepackage{subcaption}
\usepackage{booktabs} %

\usepackage{xr}

\makeatletter
\newcommand*{\addFileDependency}[1]{%
  \typeout{(#1)}
  \@addtofilelist{#1}
  \IfFileExists{#1}{}{\typeout{No file #1.}}
}
\makeatother

\usepackage{newfloat}
\usepackage{listings}
\DeclareCaptionStyle{ruled}{labelfont=normalfont,labelsep=colon,strut=off} %
\floatstyle{ruled}
\newfloat{listing}{tb}{lst}{}
\floatname{listing}{Listing}
\title{Enhancing the Antidote: Improved Pointwise Certifications \\ against Poisoning Attacks}
\author {
    Shijie Liu\textsuperscript{\rm 1},
    Andrew C. Cullen\textsuperscript{\rm 1},
    Paul Montague\textsuperscript{\rm 2},
    Sarah M. Erfani\textsuperscript{\rm 1},
    Benjamin I. P. Rubinstein\textsuperscript{\rm 1}
}
\affiliations {
    \textsuperscript{\rm 1}School of Computing and Information Systems, University of Melbourne, Melbourne, Australia\\
    \textsuperscript{\rm 2}Defence Science and Technology Group, Adelaide, Australia\\
    shijiel2@student.unimelb.edu.au, andrew.cullen@unimelb.edu.au, paul.montague@dst.defence.gov.au, sarah.erfani@unimelb.edu.au, benjamin.rubinstein@unimelb.edu.au
}

\begin{document}

\maketitle

\begin{abstract}
Poisoning attacks can disproportionately influence model behaviour by making small changes to the training corpus. While defences against specific poisoning attacks do exist, they in general do not provide any guarantees, leaving them potentially countered by novel attacks. In contrast, by examining worst-case behaviours Certified Defences make it possible to provide guarantees of the robustness of a sample against adversarial attacks modifying a finite number of training samples, known as pointwise certification. We achieve this by exploiting both Differential Privacy and the Sampled Gaussian Mechanism to ensure the invariance of prediction for each testing instance against finite numbers of poisoned examples. In doing so, our model provides guarantees of adversarial robustness that are more than twice as large as those provided by prior certifications.
\end{abstract}

\section{Introduction}
Despite the impressive performance, many modern machine learning models have been shown to be vulnerable to adversarial data perturbations~\cite{biggio_poisoning_2013, chen_targeted_2017, chakraborty_adversarial_2018}. This adversarial sensitivity is a significant concern now that machine learning models are increasingly being deployed in sensitive applications. Of particular concern are data poisoning attacks, where an adversary manipulates the training set to change the decision boundary of learned models. The risk of such attacks is heightened by the prevalence of large, user-generated datasets that are constructed without vetting. The fact that these attacks can render a model useless further underscores the need for robust defence mechanisms. Some examples of models that are vulnerable to data poisoning attacks include email spam filters and malware classifiers. These models have been shown to be susceptible to attacks that either render the model ineffective~\cite{biggio_poisoning_2013}, or that produce targeted misclassifications~\cite{chen_targeted_2017}.

The defences intrinsically counter specific poisoning attacks means that even state-of-the-art defences~\cite{carnerero-cano_regularisation_2020, paudice_detection_2018} can be vulnerable to new attacks. To circumvent this inherent dependency of defences on attacks, recent work has begun to consider the construction of guarantees of predictive invariance against bounded numbers of poisoned training examples. This is known as the \textit{certified robustness}, which is commonly achieved through the addition of calibrated noise through \textit{randomised smoothing}~\cite{lecuyer_certified_2019}. While these certifications have been successfully applied to poisoning attacks on labels and/or input features \cite{rosenfeld_certified_2020, wang_certifying_2020}, their applicability has been limited to attacks that modify training examples, rather than the more general insertion/deletion operations. On the other hand, classifiers trained with \textit{differential privacy} (DP) can be shown to be certifiably robust against poisoning attacks even against insertion/deletion operations~\cite{ma_data_2019, hong_effectiveness_2020}. However, to date, such certifications do not provide \emph{pointwise guarantees} which ensures the robustness for individual samples against a finite number of poisoned training examples. This omission still leaves a vulnerability that can be exploited by a motivated adversary to compel the model to misclassify a particular testing sample. Recent works~\cite{jia_intrinsic_2020, levine_deep_2021} leveraging bagging have achieved pointwise guarantees against poisoning attacks that allow insertion/deletion. However, some of these methods are specialized to particular learning approaches.

In this work, we establish a general framework for deriving pointwise-certifiably robust guarantees against data poisoning attacks that can influence both the label and feature sets. Such guarantees ensure that the predicted class of an individual sample are invariant to a finite number of changes to the training dataset. Prior works have leveraged DP to improve statistical properties of certification against data poisoning across a dataset. In contrast, we are the first to extend DP to certify individual samples. By producing an \emph{improved group privacy for the Sampled Gaussian Mechanism}, our new approach even yields certifications that hold for more changes to the training dataset than what had been identified by prior approaches~\cite{ma_data_2019, jia_intrinsic_2020, levine_deep_2021}. Our specific achievements can be summarized as follows:

\begin{itemize}
    \item A general framework providing \emph{pointwise-certified robustness} guarantees for models that are trained with differentially-private learners. 
    \item The framework provides a \emph{general poisoning attack defence} against insertion, deletion, and modification attacks on both the label and feature sets. The defence improves the existing differential privacy based approaches, and its efficiency is enhanced through optimised group privacy in the Sampled Gaussian Mechanism and sub-sampled training.
    \item Our defence method achieves more than double the number of poisoned examples compared to existing certified approaches as demonstrated by experiments on MNIST, Fashion-MNIST and CIFAR-$10$.
\end{itemize}

\section{Data Poisoning Attacks and Defences}

Training-time or data poisoning attacks~\cite{barreno2006can,biggio_security_2014} enable malicious users to manipulate training data and modify the learned model. The expressiveness of machine learning model families makes modern machine learning particularly susceptible to such attacks~\cite{chakraborty_adversarial_2018, goldblum_dataset_2021}. These attacks can be taxonomically classified as either label attacks, which only modify dataset labels~\cite{xiao_adversarial_2012}; features attacks, in which the training features are modified~\cite{shafahi_poison_2018}; or example attacks, such as the backdoor, which seek to influence both labels and features of the training corpus~\cite{shafahi_poison_2018}. Defending against any of these attacks is inherently complex, as their existence implies that the attacker has access to both the training architecture and dataset. Although previous works have examined attackers who solely modify the training data, our threat model assumes a more comprehensive scenario, whereby attackers have the freedom to introduce or remove samples from the training dataset, as outlined in Table~\ref{tab:related works comparison}. However, this freedom is subject to certain constraints that aim to reduce the probability of detection.

\paragraph{Threat Model.} We consider supervised multi-class classification on a training dataset $\mathcal{D}_1=\{(\x_i,y_i)\}_{i=1}^n$, where each example comprises an input instance $\x\in\mathbb{R}^m$ and label  $y_i\in\mathcal{L}=\{1,\ldots,L\}$. Consider a (possibly randomised) learner $M$ taking $\mathcal{D}_1$ to parameters $\param\in\Theta$. We refer to learned parameters and \emph{model} interchangeably. %

In this paper we consider alternate forms of inferred scores per class for randomised learners on a given input instance $\x\in\mathbb{R}^m$ as $I_l(\x, \param)$, such that $\sum_{l\in\mathcal{L}} I_l(\x, \param)=1$ and $I_l(\x, \param) \in [0,1]$, necessitating alternate choices of $I_l(\cdot)$. Let's consider a function $\y(\x, \param)$ that returns a deterministic vector of predicted class scores in $\mathbb{R}^L$, with the $i$th component denoted $y_i(\x,\param)\in\mathbb{R}$. For example, the softmax layer of a deep network outputs a score per class, these $y_i$ sit in $[0,1]$ and sum to unity.

\begin{definition}[Inference by multinomial label]\label{def:multinomial}
Define the \emph{multinomial label} inference function as 
$$
I_l
(\x, M(\mathcal{D}_1)) = \operatorname{Pr}[\arg\max_{i} y_i(\x, M(\mathcal{D}_1))=l]\enspace.
$$
Conditioned on a deterministic model $\theta$, we may make a prediction on \x as the highest $y_i$ score; however, with these predictions as random induced by $M$, we make inferences given a training dataset as the most likely prediction.
\end{definition}

\begin{definition}[Inference by probability scores]\label{def:probscores}
Define the \emph{probability scores} inference function as 
$$
I_l(\x, M(\mathcal{D}_1)) = \mathbb{E}[ y_l(\x, M(\mathcal{D}_1))]\enspace. 
$$
In other words, we consider the \y scores as random variables (due to the randomness in learner $M$) conditional on training dataset $\mathcal{D}_1$ and input instance \x. We infer the class $l$ with the largest expected score $y_l$.
\end{definition}

These two inference rules capture alternate approaches to de-randomising class predictions and offer different relative advantages in terms of robustness. We discuss this further %
in the ``Outcomes-Guaranteed Certifications" section.

The attacker is assumed to have perfect knowledge of both the dataset $\mathcal{D}_1$, learner $M$, and inference rule $I$ (\ie a white-box attacker) with unbounded computational capabilities. However, in order to minimise the likelihood of an attack being detected, it assumed that a finite number $r\in\mathbb{N}$---known as the radius---of changes to the dataset. To reflect the assumed level of access of the attacker, these changes can take the form of additions, deletions, or modifications. We consider the attacker as attempting to achieve
\begin{equation}\label{eqn:attacker_target}
    \arg\max_{l\in\mathcal{L}}  I_l(\x, M(\mathcal{D}_2)) \neq \arg\max_{l\in\mathcal{L}}  I_l(\x, M(\mathcal{D}_1))\enspace,
\end{equation}
subject to the bound
\begin{equation}
    \mathcal{B}(\mathcal{D}_1,r):=\left\{\mathcal{D}_2:\lvert \mathcal{D}_1 \ominus \mathcal{D}_2\rvert \leq r\right\}\enspace.
\end{equation}
Here $\lvert \mathcal{D}_1 \ominus \mathcal{D}_2\rvert$ measures the \emph{size} of the symmetric difference between datasets $\mathcal{D}_1$ and $\mathcal{D}_2$, or in other words, the minimum number of operations required to map $\mathcal{D}_1$ to $\mathcal{D}_2$. The objective of the defence is to achieve \emph{pointwise-certified robustness} for an individual sample $\x$ when passed through $I\circ M$. While such a threat model can be applied to any model, henceforth we will limit our consideration of randomised learners incorporating certified defences, of the form that will be described within the remainder of the paper. 

\paragraph{Certified Defences.} The concept of pointwise-certified robustness has been widely used as a testing-time defence~\cite{cohen_certified_2019, lecuyer_certified_2019}, and has recently been extended to training-time~\citet{rosenfeld_certified_2020}. Pointwise robustness is advantageous over statistical guarantees on bounds of the objective function~\cite{ma_data_2019, hong_effectiveness_2020}, as it ensures that the attacked learner will remain unchanged for finite, bounded perturbations. The nature of these perturbations, and the certification radius $r$ are intrinsically linked to the underlying threat model.

\begin{definition}[Pointwise-Certified Robustness]\label{def:certs}
A learner is said to be \emph{$r$-pointwise-certified robust} poisoning attacks, at input instance $\x$, if there exists no $\mathcal{D}_2 \in \mathcal{B}(\mathcal{D}_1,r)$ such that~\cref{eqn:attacker_target} is true. A learner $M$ is said to be \emph{$(\eta,r)$-pointwise-certified robust} if it is $r$-pointwise-certified robust with probability at least $1-\eta$ in the randomness of $M$.
\end{definition}
In other words, the prediction of the poisoned model remains the same (or the same w.h.p.), as the poisoned dataset does not alter the probabilities of labels sufficiently to change the predicted classification.

One approach for achieving pointwise certification is randomised smoothing~\cite{rosenfeld_certified_2020, weber_rab_2021}, in which a new classifier is created such that its prediction is defined as the most probable class returned by the original classifier under  calibrated input noise. While this noise is often applied directly to the input samples, it has been shown that model bagging can also generate output randomness in a fashion that allows for certifications against data poisoning attacks~\cite{jia_intrinsic_2020, chen_framework_2020, levine_deep_2021}.

\section{Outcomes Guarantee}

By exploiting both DP and the Sampled Gaussian Mechanism (SGM), our certification framework empirically improves pointwise certifications against data poisoning. Such certificates can be used to quantify the confidence in a sample's prediction, in the face of potential dataset manipulation. 
To support our enhancements, we will first define some key properties of DP, then propose the outcomes guarantee that generalises to most DP mechanisms, and finally introduce the SGM with improved group privacy.

\begin{table*}[ht]
\centering
\begin{tabular}{p{0.4\linewidth}p{0.1\linewidth}p{0.1\linewidth}p{0.1\linewidth}p{0.1\linewidth}}
\toprule
                                                                     & \multicolumn{2}{c}{\textit{Training-time threat model}}       & \multicolumn{2}{c}{\textit{Testing-time certification}}                          \\[0em] \cmidrule(lr){2-5}
                                                                     & \multicolumn{1}{p{0.1\linewidth}}{Modification} & Addition/ Deletion & \multicolumn{1}{p{0.1\linewidth}}{Statistical certification} & Pointwise certification \\ \midrule
Statistical DP~\cite{ma_data_2019}                                   & \multicolumn{1}{l}{\cmark}   & \cmark        & \multicolumn{1}{l}{\cmark}                & \xmark                \\ \midrule
Randomized smoothing~\cite{rosenfeld_certified_2020, weber_rab_2021} & \multicolumn{1}{l}{\cmark}   & \xmark          & \multicolumn{1}{l}{\cmark}                & \cmark              \\ \midrule
Bagging~\cite{jia_intrinsic_2020, levine_deep_2021}                  & \multicolumn{1}{l}{\cmark}   & \cmark        & \multicolumn{1}{l}{\cmark}                & \cmark              \\ \midrule
This Paper                                                           & \multicolumn{1}{l}{\cmark}   & \cmark        & \multicolumn{1}{l}{\cmark}                & \cmark              \\ \bottomrule

\end{tabular}
\caption{A summary of different approaches of certified defence against data poisoning attacks. We investigate them from two perspectives. The training-time threat model: whether it permits the more general addition/deletion of training samples or only modification. The testing-time certification: whether it provides the more strict pointwise certification for each test sample or only statistical certification over all test samples.}
\label{tab:related works comparison}
\end{table*}

\paragraph{Differential Privacy.} A framework~\cite{dwork2006calibrating, abadi_deep_2016, friedman2010data} quantifies the privacy loss due to releasing aggregate statistics or trained models on sensitive data.  As a versatile notation of smoothness to input perturbations, DP has successfully been used as the basis of several certification regimes.

\begin{definition}[Approximate-DP]
A randomised function $M$ is said to be $(\epsilon, \delta)$-approximate DP (ADP) if for all datasets $\mathcal{D}_1$ and $\mathcal{D}_2$ for which $\mathcal{D}_2 \in \mathcal{B}(\mathcal{D}_1,1)$, and for all measurable output sets $\mathcal{S} \subseteq \operatorname{Range}(M)$: %
\begin{equation}
\label{def:dp equation}
    \operatorname{Pr}[M(\mathcal{D}_1) \in \mathcal{S}] \leq \operatorname{exp}(\epsilon) \operatorname{Pr}[M(\mathcal{D}_2) \in \mathcal{S}] + \delta\enspace,
\end{equation}
where $\epsilon > 0$ and $\delta \in [0,1)$ are chosen parameters.
\end{definition}

Smaller values of the privacy budget $\epsilon$ tighten the (multiplicative) influence of a participant joining dataset $\mathcal{D}_{2}$ to form $\mathcal{D}_{1}$, bounding the probability of any downstream privacy release. The confidence parameter $\delta$ then relaxes this guarantee, such that no bound on the privacy loss is provided for low-probability events.

To bound the residual risk from ADP, R\'{e}nyi-DP was introduced by \citet{mironov_renyi_2017}. R\'{e}nyi-DP quantifies privacy through sequences of function composition, as required when iteratively training a deep net on sensitive data, for example. As we shall see in this paper, this tighter analysis leads to improved certifications in practice.

\begin{definition}[Rényi divergence]
Let $P$ and $Q$ be two distributions on $\mathcal{X}$ defined over the same probability space, and let $p$ and $q$ be their respective densities. The Rényi divergence of finite order $\alpha \neq 1$ between $P$ and $Q$ is defined as
\begin{equation}
    \mathrm{D}_{\alpha}(P \| Q) \triangleq \frac{1}{\alpha-1} \ln \int_{\mathcal{X}} q(x)\left(\frac{p(x)}{q(x)}\right)^{\alpha} \mathrm{d} x\enspace.
\end{equation}
\end{definition}

\begin{definition}[Rényi Differential Privacy]
A randomised function $M$ preserves $(\alpha, \epsilon)$-Rényi-DP, with $\alpha>1, \epsilon>0$, if for all datasets $\mathcal{D}_1$ and $\mathcal{D}_2 \in \mathcal{B}(\mathcal{D}_1,1)$:
\begin{equation}
\label{def:rdp equation}
    \mathrm{D}_{\alpha}\left(M(\mathcal{D}_1) \| M\left(\mathcal{D}_{2}\right)\right) \leq \varepsilon \enspace.
\end{equation}
\end{definition}

\begin{definition}[Outcomes guarantee]
\label{def:outcomes guarantee}
Let $\mathcal{K}$ be a set of strictly monotonic functions on the reals, and $r$ a natural number.
A randomised function $M$ is said to preserve a \emph{$(\mathcal{K},r)$-outcomes guarantee} if for any $K\in\mathcal{K}$ such that for all datasets $\mathcal{D}_1$ and $\mathcal{D}_2\in\mathcal{B}(\mathcal{D}_1,r)$,
\begin{equation}
\label{equ:outcomes guarantee}
    \operatorname{Pr}[M(\mathcal{D}_1) \in \mathcal{S}] \leq \operatorname{K} (\operatorname{Pr}[M(\mathcal{D}_2) \in \mathcal{S}])\enspace.
\end{equation}

\end{definition}

Both ADP and RDP are generalised as specific cases of the outcome guarantee with, respectively, 
\begin{align}
\label{equ:dp outcomes guarantee}
    \mathcal{K}_{\epsilon, \delta}(x)&=\operatorname{exp}(\epsilon) x + \delta\\
    \mathcal{K}_{\epsilon, \alpha}(x)&=(\operatorname{exp}(\epsilon) x)^{\frac{\alpha-1}{\alpha}}\enspace.  \label{equ:outcomes guarantee rdp}
\end{align}
The RDP's family is obtained by applying Hölder's inequality to the integral of the density function in the Rényi divergence~\cite{mironov_renyi_2017}. 

This definition formalises a discussion on ``bad-outcomes guarantee'' due to \citet{mironov_renyi_2017}. With this definition, we are able to generalise our certification framework to the essential structure across variations of differential privacy. Note this definition incorporates \emph{group privacy}~\cite{dwork2006calibrating}: extending DP to pairs of datasets that differ in up to $r$ data-points $\mathcal{D}_1 \in \mathcal{B}(\mathcal{D}_2,r)$.

Our framework also relies upon the \emph{post-processing} property~\cite{dwork2006calibrating} of DP: any computation applied to the output of a DP algorithm retains the same DP guarantee. This property, which simplifies the DP analysis in multi-layered models acting on a DP-preserving input, holds for any ADP, RDP, or indeed outcome-guaranteed mechanism.

\eat{In \citet{mironov_renyi_2017} some proofs of bad-outcomes guarantee hold for both ADP and RDP. As such, where relevant, we state results in terms of the more general outcomes guarantee, while specifying the necessary assumptions on the bounding functions $\mathcal{K}$, thus highlighting the generality of results. %

In the case of ADP the outcomes guarantee is explicitly shown in Equation~\eqref{def:dp equation} with the bound function parameterised by $\epsilon$ and $\delta$ as
\begin{equation}
\label{equ:dp outcomes guarantee}
    \mathcal{K}_{\epsilon, \delta}(x)=\operatorname{exp}(\epsilon) x + \delta \enspace.
\end{equation}
As for Rényi-DP, the outcomes guarantee can be obtained by applying Hölder's inequality to the integral of the density function in the Rényi divergence~\cite{mironov_renyi_2017}. The bound function is parameterised by $\epsilon$ and $\alpha$ via Equation~\eqref{def:rdp equation} as
\begin{equation}
\label{equ:outcomes guarantee rdp}
    \mathcal{K}_{\epsilon, \alpha}(x)=(\operatorname{exp}(\epsilon) x)^{\frac{\alpha-1}{\alpha}}\enspace. 
\end{equation}
}

\paragraph{Sampled Gaussian Mechanism with Improved Group Privacy.} While many DP mechanisms have been proposed and widely studied for machine learning~\cite{abadi_deep_2016, mironov_renyi_dp_2019}, they typically rely upon the addition of noise to training samples. In contrast, the Sampled Gaussian Mechanism (SGM)~\cite{mironov_renyi_dp_2019} adds randomness both through the injection of noise and the sub-sampling process. Each element of the training batch is sampled without replacement with uniform probability $q$ from the training dataset, while each weight update step also introduces additional additive gaussian noise to the gradients. When applied to a model $M$, the SGM has been shown~\cite{mironov_renyi_dp_2019} to preserve $(\alpha, \epsilon)$-Renyi-DP, where $\epsilon$ is determined by the parameters $(\alpha, M, q, \sigma)$. We denote the computation steps of $\epsilon$ in SGM as function $\operatorname{SG}$ such that $\epsilon = \operatorname{SG}(\alpha, M, q, \sigma)$.

However, this guarantee fails to exploit the advantages given by Rényi-DP group privacy under the SGM. As such, by constructing our group privacy in a manner specific to the SGM, we are able to produce \emph{tighter bounds} than prior works~\cite{mironov_renyi_2017}, producing the following pointwise guarantee of certification.

\begin{theorem}[Improved Rényi-DP group privacy under the SGM]
\label{the:Improved rdp group privacy in SGM}
    If a randomised function $M$ obtained by SGM with sample ratio $q$ and noise level $\sigma$ achieves $(\alpha, \operatorname{SG}(\alpha, M, q, \sigma))$-Rényi-DP for all datasets $\mathcal{D}_1$ and $\mathcal{D}_2 \in \mathcal{B}(\mathcal{D}_1,1)$, then for all datasets $\mathcal{D}_3 \in \mathcal{B}(\mathcal{D}_1,r)$
    \begin{equation}
        \mathrm{D}_{\alpha}\left(M(\mathcal{D}_1) \| M\left(\mathcal{D}_{3}\right)\right) \leq \operatorname{SG}(\alpha, M, 1-(1-q)^{r}, \sigma)\enspace.
    \end{equation}
    \begin{proof}
        (Here we provide a proof sketch, the detailed proof is available in Appendix A.2.) In the work~\cite{mironov_renyi_dp_2019}, they proposed calculating the amount of Rényi-DP obtained from SGM in their Theorem $4$. We extend it from ``adjacent datasets'' to ``datasets that differ in up to $r$ examples''. We consider the datasets $S$ and $S^{\prime}=S \cup\{x_1, x_2,..., x_r\}$, and calculate the mixing of distributions of taking a random subset of $S'$ by the SGM $\mathcal{M}$ where each element of $S'$ is independently placed with probability $q$ as
        $$
        \begin{aligned}
            &\mathcal{M}\left(S^{\prime}\right)=\sum_{T} p_{T}\left(\sum_{k=0}^{r}{\binom{r}{k}}
            q^{k}(1-q)^{r-k}\mathcal{N}\left(\mu, \sigma^{2} \mathbb{I}^{d}\right)\right)\\
            &V \subseteq \{x_1, x_2,..., x_r\} \qquad k = \|V\| \enspace.%
        \end{aligned}
        $$ We can complete the proof by replacing the original $\mathcal{M}(S')$ by the above distribution, and by following the remainder of the paper.
    \end{proof}
\end{theorem}

\section{Outcomes-Guaranteed Certifications}
\label{sec:Differential Privacy Certified Defence Framework}
While pointwise-certified robustness guarantees can be applied to the output of any model, within this work we highlight both multinomial outputs and scored outputs. 
\begin{lemma}[Pointwise outcomes guarantee]
\label{lem:general DP output bounds}
Consider a randomised learner $M$ that preserves a $(\mathcal{K},r)$-outcome guarantee,
and an arbitrary (possibly randomised) inference function $I$ mapping learned parameters and the input instance to an inferred score.
Then for any $K\in\mathcal{K}$ such that, for any input instance $\x$, label $l\in\mathcal{L}$, training datasets $\mathcal{D}_1$ and $\mathcal{D}_2\in\mathcal{B}(\mathcal{D}_1,r)$,
\begin{align*}%
    I_l\left(\x, M\left(\mathcal{D}_1\right)\right) & \leq \operatorname{K}\left( I_l\left(\x, M\left(\mathcal{D}_2\right)\right)\right) \\
    I_l\left(\x, M\left(\mathcal{D}_1\right)\right) & \geq \operatorname{K}^{-1}\left( I_l\left(\x, M\left(\mathcal{D}_2\right)\right)\right)\enspace. 
\end{align*}
\end{lemma}
\begin{proof}
\eat{Here $M_{DP}(\mathcal{D})$ is regarded as the final parameters $\theta \in \Theta$ of the model which satisfies the DP outcomes guarantee. Since the inference process can be interpreted as a computation over the output of $\mathcal{M_{DP}}(\mathcal{D})$ (final parameters $\theta$) and the testing sample $\x$, the output of $M_{DP}(\mathcal{D}, \x)$ still satisfies the identical DP guarantee with regards to the training dataset $\mathcal{D}$ by the post-processing property. Thus the outcomes guarantee implies
\begin{equation}
\label{equ:trans outcomes guarantee}
    \operatorname{Pr}[M_{DP}(\mathcal{D}_1, \x) \in \mathcal{S}] \leq \operatorname{K}_{r}(\operatorname{Pr}[M_{DP}(\mathcal{D}_2, \x) \in \mathcal{S}])
\end{equation}
for any $\mathcal{D}_{2}$ and $\mathcal{D}_1 \in \mathcal{B}(\mathcal{D}_2,r)$. 
}
In the case of multinomial outputs, the first inequality follows from the post-processing property: the composition $I\circ M$ preserves the same outcome guarantee. The second inequality follows by symmetry in the roles of $\mathcal{D}_1,\mathcal{D}_2$ and by $K$ being strictly increasing. To admit scored outputs, the probabilities $\operatorname{Pr}[M(\mathcal{D}) \in \mathcal{S}]$ in $(\mathcal{K},r)$-outcome guarantee need to be converted into expected values $\mathbb{E}[M(\mathcal{D})]$. To that end, the integral over the right-tail distribution function of the probabilities in Definition \ref{def:outcomes guarantee} are taken. 
\eat{
Let us now consider the label range space $\mathcal{L}$ of $M_{DP}(\mathcal{D}, \x)$ as partitioned into singletons $\{l_=i\} \subseteq \mathcal{L}$. Then by setting $\mathcal{S}=\{l_i\}$, $\mathcal{D}_1=\mathcal{D}'$, and $\mathcal{D}_2=\mathcal{D}$ in Equation \eqref{equ:trans outcomes guarantee}, it follows that an upper bound can be constructed such that: 
\begin{equation}
\begin{aligned}
    & \operatorname{Pr}[M_{DP}(\mathcal{D}', \x) \in \{l_i\}] \leq \operatorname{K}_{r}( \operatorname{Pr}\left[M_{DP}\left(\mathcal{D}, \x\right) \in \{l_i\}\right]) \\
    \rightarrow & \operatorname{Pr}[M_{DP}(\mathcal{D}', \x)=l_i] \leq \operatorname{K}_{r}( \operatorname{Pr}\left[M_{DP}\left(\mathcal{D}, \x\right)=l_i\right])
\end{aligned}
\end{equation}
for any $\mathcal{D}$ and  $\mathcal{D'} \in \mathcal{B}(\mathcal{D},r)$. Due to the symmetry between $\mathcal{D}$ and $\mathcal{D}'$, the lower bound of
\begin{equation}
    \operatorname{Pr}[M_{DP}(\mathcal{D}', \x)=l_i] \geq \operatorname{K}^{-1}_{r}( \operatorname{Pr}\left[M_{DP}\left(\mathcal{D}, \x\right)=l_i\right])
\end{equation}
can be constructed trivially by way of symmetry.
}
The expected value $(\mathcal{K},r)$-outcome guarantee of $(\epsilon, \delta)$-ADP and $(\alpha, \epsilon)$-R\'{e}nyi-DP can be shown to take the same form of \cref{equ:dp outcomes guarantee} and \cref{equ:outcomes guarantee rdp} by~\citet{lecuyer_certified_2019} and Hölder's inequality (as detailed in Appendix A.3) respectively.
\end{proof}

The main result of this section establishes conditions under which a DP learner %
provides pointwise-certified robustness against general poisoning attacks up to size $r$. 

\begin{theorem}[Pointwise-certified robustness by outcomes guarantee]
\label{the:dp pointwise-certified robustness guarantee}
Consider a training dataset $\mathcal{D}$, an input instance $\x$, and a randomised learner $M$ that preserves a $(\mathcal{K},r)$-outcomes guarantee. 
Let 
$$l_1 = \arg\max_{l\in\mathcal{L}} I_l(\x, M(\mathcal{D}))$$
denote the label predicted on $\x, \mathcal{D}$ under multinomial interpretation of Definition~\ref{def:multinomial}. 
If there exist $\operatorname{K}_{upper}, \operatorname{K}_{lower} \in \mathcal{K}$ such that 
\begin{align*}
    \operatorname{K}_{lower}^{-1} (I_{l_1}(\x, M(\mathcal{D}))) & > \\
    \max_{l \in \mathcal{L} \setminus \{l_1\}} \operatorname{K}_{upper}( I_l(\x, M(\mathcal{D}))) & \phantom{\ >} \enspace,
\end{align*}
then $I\circ M$ is pointwise-certified robust to radius $r$ about dataset $\mathcal{D}$ at input $\x$ (see Definition~\ref{def:certs}). %
\end{theorem}
The proof can be found in Appendix A.1.

\subsection{Algorithmic Implementation}
\label{sec:practical algorithm}

The very nature of data poisoning attacks intrinsically requires pointwise certifications to incorporate modifications to both the training and testing processes. The remainder of this section illustrates the steps required to produce a prediction and certification pair $(l, r)$ for a test time sample $\x$ in the testing dataset $\mathcal{D}_e$, the details of which are further elaborated over the \cref{alg:DP defence algorithm}.

\paragraph{Training.} Any certification using the aforementioned DP based processes inherently requires the model $M_{DP}$ to be randomised. The SGM achieves the requisite randomness for DP via sub-sampling and injecting noise. The randomised model $M_{DP}$ is instanced such that $(\hat{M}_{DP_{1}}, \hat{M}_{DP_{2}},..., \hat{M}_{DP_{P}})$. As each instance is a model with an identical training process, the training of such is an embarrassingly parallel process, a fact that can be leveraged to improve training efficiency. Further efficiencies for larger datasets can be found by incorporating training over subset $D_{sub} \subseteq D$. Under the SGM, the total privacy cost with regards to $\mathcal{D}$ is calculated by accumulating the privacy cost of each update with a subsample from $\mathcal{D}$. Therefore, we can analogously compute the privacy cost of using a sub-training dataset $D_{sub} \subseteq D$ with regards to the entire training dataset $\mathcal{D}$ by reducing the number of updates under the SGM. 

Rather than exploiting the SGM, an alternate approach is to construct sub-training datasets across a set of model instances through bagging. Taking such an approach allows the model instance to work on a subset solely without knowing the entire training dataset. The privacy gains can be quantified by way of DP amplification~\cite{balle_privacy_2018}. However, both SGM and bagging yield a level of privacy can then be translated into certifications of size $r$ by \cref{the:Improved rdp group privacy in SGM} by deriving the set of outcomes guarantee bound functions $\mathcal{K}$ through \cref{alg:DP defence algorithm}.

\begin{algorithm*}[tb]
   \caption{Certifiably Robust Differentially Private Defence Algorithm}
   \label{alg:DP defence algorithm}
\begin{algorithmic}
  \STATE {\bfseries Input:} model $M_{DP}$, training dataset $\mathcal{D}$, testing dataset $\mathcal{D}_e$, number of instances $P$, confidence interval $1-\eta$
  \STATE $(\hat{M}_{DP_{1}}, \hat{M}_{DP_{2}},..., \hat{M}_{DP_{P}}) \leftarrow \textsc{DifferentialPrivateTrain}(M, P)$ \COMMENT{Train DP model instances}
  \FOR{\textbf{each} $x_i \in \mathcal{D}_e$} 
  \STATE counts[$j$] $\leftarrow \sum_{o=1}^{P} \mathbb{I}\left(\hat{M}_{DP_{o}}\left(\mathbf{x}_{i}\right)=j\right), j \in \mathcal{L}$ \COMMENT{Count votes for each label}
  \STATE $l_{1i}, l_{2i} \leftarrow$ top two indices in counts 
  \STATE $\underline{p_{l_{1i}}}, \overline{p_{l_{2i}}} \leftarrow \textsc{SimuEM}(\text{counts}, 1-\eta)$ \COMMENT{Calculate the corresponding lower and upper bounds}
  \IF{ $\underline{p_{l_{1i}}} \geq \overline{p_{l_{2i}}}$}
  \STATE $r_i \leftarrow \textsc{BinarySearchForCertifiedRadius}(\underline{p_{l_{1i}}}, \overline{p_{l_{2i}}}, \|D\|)$ \COMMENT{Calculate certified radius if possible}
  \ELSE
  \STATE $r_i \leftarrow \textsc{ABSTAIN}$ \COMMENT{Unable to certify}
  \ENDIF
  \ENDFOR
  \STATE {\bfseries Output:} $l_{11}, l_{12},.., l_{1e}$ and $r_1, r_2,.., r_e$
\end{algorithmic}
\end{algorithm*}

\paragraph{Certification.} %
In general, the certification involves estimating the upper and lower bounds of inferred scores for each label and searching for the maximum radius that satisfies \cref{the:dp pointwise-certified robustness guarantee}. The \emph{multinomial label} and \emph{probability scores} require similar but slightly different treatments. For the former, each testing sample $\x_i$ is passed through the set of model instances $(\hat{M}_{DP_{1}}, \hat{M}_{DP_{2}},..., \hat{M}_{DP_{P}})$. From this, the top-$2$ most frequent labels are selected and labelled as $l_{1i}$ and $l_{2i}$. Uncertainties from sampling are then quantified through the lower and upper confidence bounds of $\operatorname{Pr}[M_{DP}(\x_i, \mathcal{D}) = l_{1i}]$  and $\operatorname{Pr}[M_{DP}(\x_i, \mathcal{D}) = l_{2i}]$, which are constructed to a confidence level $1-\eta$ by the $\textsc{SimuEM}$ method of~\citet{jia_intrinsic_2020}, yielding $\underline{p_{l_{1i}}}$ and $\overline{p_{l_{2i}}}$ respectively. 

\cref{alg:DP defence algorithm} demonstrates that a binary search can then be used to identify the maximum certified radius $r_i$ of the optimisation problem in \cref{the:dp pointwise-certified robustness guarantee}, subject to 
\begin{equation}
\begin{aligned}
&\operatorname{K}_{lower}^{-1} (\operatorname{Pr}\left[M_{DP}(\x_i, \mathcal{D})=l_{1i}\right]) = \operatorname{K}_{lower}^{-1} (\underline{p_{l_{1i}}}) \\ 
&\max_{l_{ji} \in \mathcal{L} \setminus \{l_{1i}\}} \operatorname{K}_{upper}( \operatorname{Pr}\left[M_{DP}(\x_i, \mathcal{D})=l_{ji}\right]) = \operatorname{K}_{upper}(\overline{p_{l_{2i}}})
\end{aligned}
\end{equation}
Here the bound functions $\operatorname{K}_{upper}, \operatorname{K}_{lower} \in \mathcal{K}$ being derived by \cref{lem:general DP output bounds}. The outputs for a testing sample $\x_i$ are the predicted label $l_{1i}$ with certified radius $r_i$. 

The process for the \emph{probability scores} case is similar but involves collecting the probability scores from each model instance and computing the confidence interval for the expected values $\mathbb{E}[y_{l_i}(\x_i, M_{DP}(\mathcal{D})]$ via Hoeffding's inequality~\cite{hoeffding_probability_1963} or empirical Bernstein bounds~\cite{maurer_empirical_2009}. 

\newcommand{\figueWidFrac}{.69}
\begin{figure*}
  \begin{subfigure}[b]{0.5\linewidth}
    \centering
    \includegraphics[width=\figueWidFrac\linewidth]{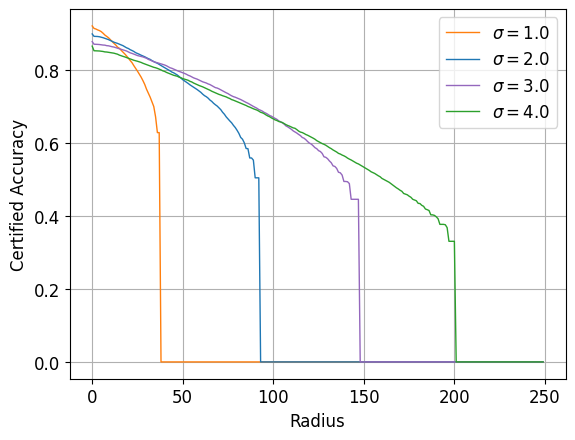} 
    \caption{MNIST (LeNet-5), performance against different $\sigma$.} 
  \end{subfigure}%
  \begin{subfigure}[b]{0.5\linewidth}
    \centering
    \includegraphics[width=\figueWidFrac\linewidth]{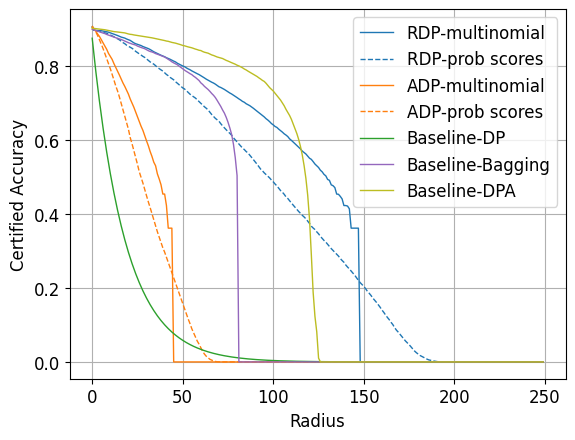} 
    \caption{MNIST (LeNet-5), comparative performance at $\sigma=3.0$.}
  \end{subfigure} 
  \begin{subfigure}[b]{0.5\linewidth}
    \centering
    \includegraphics[width=\figueWidFrac\linewidth]{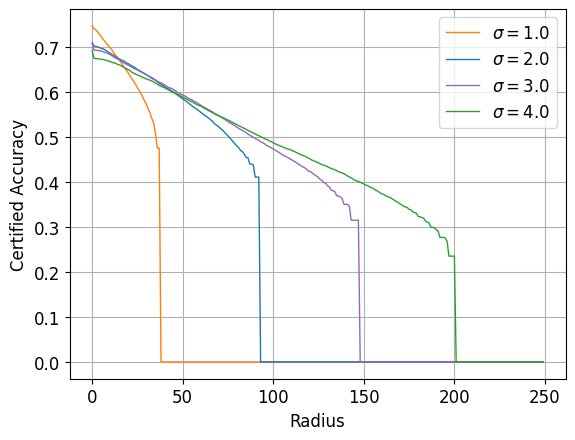} 
    \caption{Fashion-MNIST (LeNet-5), performance against different $\sigma$.} 
  \end{subfigure}%
  \begin{subfigure}[b]{0.5\linewidth}
    \centering
    \includegraphics[width=\figueWidFrac\linewidth]{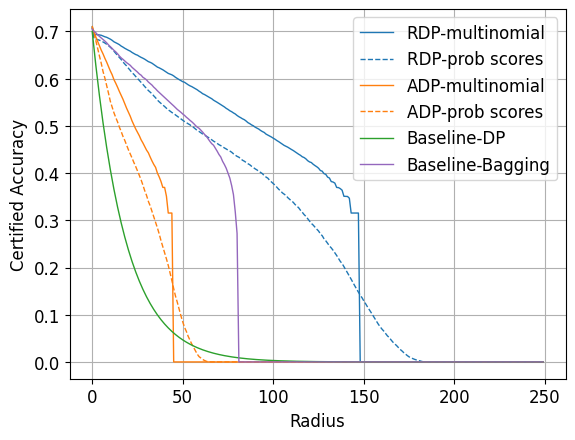} 
    \caption{Fashion-MNIST (LeNet-5), comparative performance at $\sigma=3.0$.}
  \end{subfigure} 
  \begin{subfigure}[b]{0.5\linewidth}
    \centering
    \includegraphics[width=\figueWidFrac\linewidth]{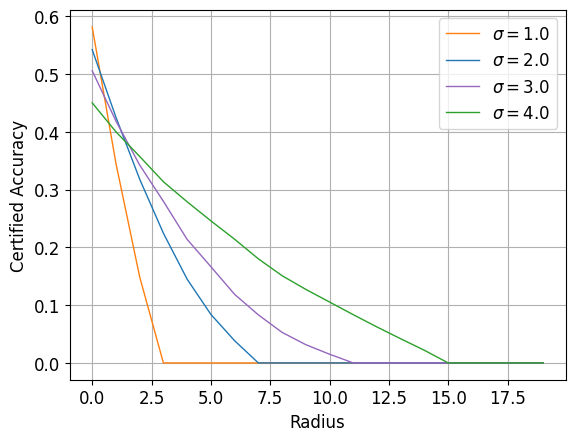} 
    \caption{CIFAR-$10$ (Opa-tut), performance against different $\sigma$.} 
  \end{subfigure}%
  \begin{subfigure}[b]{0.5\linewidth}
    \centering
    \includegraphics[width=\figueWidFrac\linewidth]{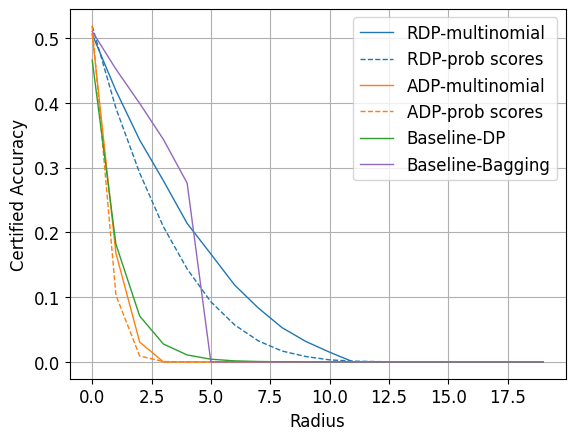} 
    \caption{CIFAR-$10$ (Opa-tut), comparative performance at $\sigma=3.0$.} 
  \end{subfigure}
  \begin{subfigure}[b]{0.5\linewidth}
    \centering
    \includegraphics[width=\figueWidFrac\linewidth]{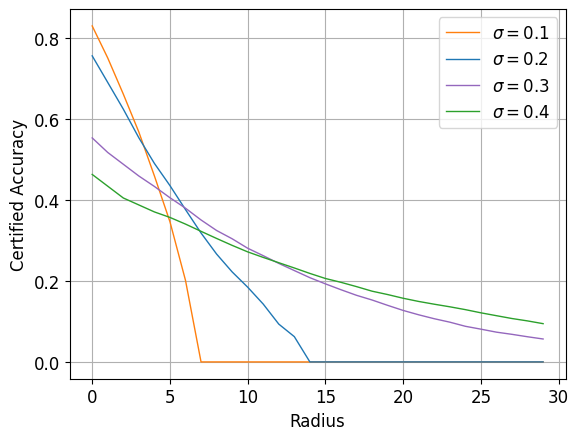} 
    \caption{CIFAR-$10$ (ResNet-18), performance against different $\sigma$.} 
  \end{subfigure} 
  \begin{subfigure}[b]{0.5\linewidth}
    \centering
    \includegraphics[width=\figueWidFrac\linewidth]{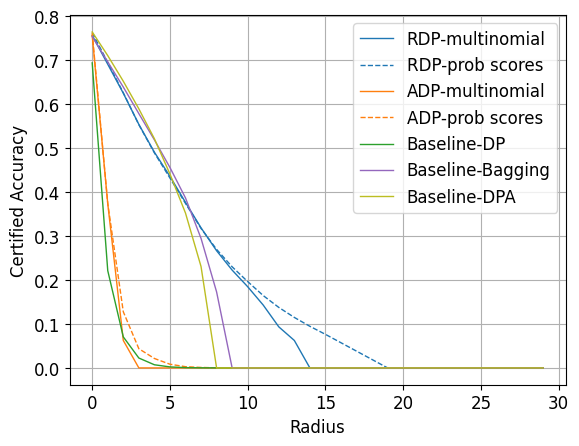}
    \caption{CIFAR-$10$ (ResNet-18), comparative performance at $\sigma=0.2$.} 
  \end{subfigure} 
  \caption{The left column contains certified accuracy plots for the method RDP-multinomial against different noise levels ($\sigma$); the right column contains certified accuracy plots for comparisons against variants and baselines. In the plots, the X-axis is radius $r$ (symmetric difference) while the Y-axis is the corresponding certified accuracy $CA_r$ at radius $r$.}
  \label{fig:consolidated}
\end{figure*}

\section{Experiments}
To verify the effectiveness of our proposed pointwise-certified defence, we conducted experiments across MNIST, Fashion-MNIST, and CIFAR-$10$ for varying levels of added noise $\sigma$. %
For MNIST and Fashion-MNIST, training occurred using the LeNet-5 architecture~\cite{lecun_gradient_based_1998}, with class probabilities/expectations estimated based upon $1000$ model instances trained on the entire dataset. In contrast, CIFAR-$10$ was trained upon the example model from Opacus tutorial~\cite{opacus} (Opa-tut) with rather simple architecture, and more complex ResNet-18~\cite{he_deep_residual_2015} for comprehensive evaluation. Both were estimated based upon $500$ instances trained on sub-datasets of size $10000$. 

Across all experiments adjust the sample ratio $q$ to have a batch size of $128$, with training conducted using ADAM with a learning rate of $0.01$ optimising the Cross-Entropy loss. The clip size $C$ is fine-tuned for each experiment (around $1.0$ on MNIST, $25.0$ on CIFAR-10). In each case, uncertainties were estimated for a confidence interval suitable for $\eta=0.001$. All experiments were conducted in Pytorch using a single NVIDIA RTX $2080$ Ti GPU with $11$ GB of GPU RAM. %

To quantify performance the proportion of samples correctly predicted with a certification of at least $r$ was used, henceforth known as the \emph{certified accuracy}. This quantity takes the form
\begin{equation}
    C A_{r}=\frac{\sum_{\x_{i} \in \mathcal{D}_{e}} \mathbb{I}\left(l_{i}=y_{i}\right) \cdot \mathbb{I}\left(r_{i} \geq r\right)}{\left|\mathcal{D}_{e}\right|}\enspace,
\end{equation}
where $\x_i$ and $y_i$ are the input instances and corresponding ground truth labels for a testing sample, and $l_i$, and $r_i$ are the predicted label and corresponding certified radius returned by the defence model. We also investigate the median and maximum value of certification achieved among all samples.

We further divide our experiments into four different frameworks. These are ADP with either multinomial labels (ADP-multinomial) or probability scores (ADP-prob-scores) output, and then Rényi-DP with either multinomial labels (RDP-multinomial) or probability scores (RDP-prob-scores) output. In each case, \cref{the:dp pointwise-certified robustness guarantee} is employed to generate a guaranteed certificate of defence to data poisoning attacks. 

To validate the efficacy of our technique, these results are considered against prior works, specifically the DP-based defence method of~\citet{ma_data_2019} (Baseline-DP), the bagging-based defence of~\citet{jia_intrinsic_2020, chen_framework_2020} (Baseline-Bagging) and deterministic Deep Partition Aggregation (DPA) method of ~\cite{levine_deep_2021} (Baseline-DPA). Of these, conceptual similarities between our work and DP-baseline allow both techniques to be compared while utilising the same trained models. However, it must be noted that \citet{ma_data_2019} bound the DP-baseline in terms of statistically certified accuracy which is calculated as the lower bound of expected accuracy with confidence level $1-\eta$ among obtained model instances. As for Bagging-baseline, it provides the same pointwise-certified defence as we do. Hence, by letting the number of base classifiers equal the number of model instances and adjusting the size of sub-training datasets, we force the Bagging-baseline to have the same certified accuracy at radius $r=0$. The DPA method has significant differences between their underlying assumptions and ours. The DPA only applies to the models that are \emph{deterministic}, which means for a given training dataset the parameters in the resulting model should always be the same. This approach requires specific model architectures and a deterministic training process while our method applies to more general situations. Compared with standard training approaches, the extra step involved in incorporating SGM introduces a negligible difference in training time. Note the change in the relative performance of Baseline-Bagging and Baseline-DPA from the original papers are the product of different model architectures. We ensure all methods apply the same model architecture for fair comparisons (Appendix A.4).

\cref{fig:consolidated} demonstrates that our method consistently provides a more robust certified defence, across the full suite of experiments. In the case of MNIST and Fashion-MNIST, for a given radius, RDP-multinomial is capable of providing the highest certified accuracy in most cases, which means more testing samples are certified to be correctly predicted within this radius. For example, in the experiments on Fashion-MNIST, RDP-multinomial achieves $52.21\%$ certified accuracy at radius $r=80$, whereas the other baselines only achieve at most $27.23\%$ certified accuracy. Additionally, our method can generate the largest certification as shown, which provides a better defence for the confident testing samples. As illustrated in the experiments on CIFAR-$10$ for both Opa-tut and ResNet-18 models, RDP-prob-scores outperform the other baselines with regard to the largest certified radius by doubling the size. Based upon these results, when considering Fashion-MNIST our method achieves a $56\%$ and $130\%$ improvement in the median and maximum value respectively when compared to Baseline-Bagging (further details of this can be found in Appendix A.6).

As the bound functions are the same in both multinomial and probability scores methods, the difference between them can be directly attributed to the differences in how these techniques construct their upper and lower bounds. As indicated in \cref{the:dp pointwise-certified robustness guarantee}, the larger the gap between the lower and upper bounds, the larger radius it can certify. Intuitively, if the defence model is confident with the predicted label of an easy testing sample, then this sample should be more resilient to poisoned examples in the training dataset. In the multinomial method, the uncertainty within each model instance is ignored by selecting a single label, while the uncertainty remains in the probability scores method. As a consequence of this, the multinomial method provides a higher radius for moderately confident examples but the probability scores method is able to certify a larger radius for the very confident ones. Further improvements can be found in the application of Renyi-DP, relative to Approximate-DP, due to the former providing a more precise accounting of model privacy. This in turn allows tighter bounds to be constructed, with performance further enhanced by way of \cref{the:Improved rdp group privacy in SGM}.%

The influence of the magnitude of injected noise $\sigma$ is shown in the left-hand column of \cref{fig:consolidated}. These results broadly align with previous works, in that adding more noise can produce larger robustness guarantees (larger certified radius), at the cost of decreased predictive performance upon un-attacked datasets ($r=0$). The increase of semantic-complexity of the dataset also limits the tolerance of the noise. It is also important to note that the sample rate ($q \in (0, 1]$) and robustness are negatively correlated, as increasing the  sample rate requires that more training examples are utilised in constructing the output, which provides weaker privacy guarantees. Therefore, a grid search is usually required to find the best combination of parameters ($\sigma$, $q$, clip size).    

\paragraph{Limitations and Future Directions}
The nature of the SGM inherently requires a significant allocation of computational resources, due to the need to train multiple models from scratch in parallel. While improvements in these resource demands may be possible, at this stage any direct application of this work would likely be restricted to systems that are considered particularly sensitive to adversarial behaviours. We also note that while this work improves upon the achievable bounds for certification by exploiting RDP in the context of the SGM, further gains may be possible by extending these proofs to Approximate DP via the conversion from RDP to ADP~\cite{balle_hypothesis_2019}.

\section{Conclusion}
By carefully exploiting both DP, SGM, and bagging, this work presents a mechanism for tightening guarantees of pointwise-certified robustness relative to prior implementations. This is made possible by calculating group privacy directly from the SGM. When compared to the current state-of-the-art, our technique can produce a more than $50\%$ improvement in the median certification.

\section*{Acknowledgements}

This research was undertaken using the LIEF HPC-GPGPU Facility hosted at the University of Melbourne. This Facility was established with the assistance of LIEF Grant LE170100200. This work was also  supported  in  part  by  the  Australian  Department  of  Defence  Next  Generation  Technologies  Fund, as part of the CSIRO/Data61 CRP AMLC project. Sarah Erfani is in part supported by the Australian Research Council (ARC) Discovery Early Career Researcher Award (DECRA) DE220100680.

\bibliography{references_offline.bib}


\onecolumn
\appendix
\section{Appendix}
\subsection{Proof of \cref{the:dp pointwise-certified robustness guarantee}}
\label{apd:proof of main theorem}
\begin{proof}
Our goal is equivalent to proving that the probability of predicting label $l_1$ by the poisoned model is larger than for any other labels, \ie 
\begin{equation}
    \begin{aligned}
    &\forall \mathcal{D'} \in \mathcal{B}(\mathcal{D},r), \\
    &\operatorname{Pr}[I(\x,  M(\mathcal{D'})) = l_1] > \max_{l_i \in \mathcal{L} \setminus \{l_1\}} \operatorname{Pr}[I(\x,  M(\mathcal{D'})) = l_i]\enspace. 
\end{aligned}
\end{equation}

Given $M$ preserves a $(\mathcal{K},r)$-outcome guarantee, then by Lemma \ref{lem:general DP output bounds}, it is possible to derive the lower bound of $\operatorname{Pr}[I(\x,  M(\mathcal{D'})) = l_1]$ and upper bounds of $\max_{l_i \in \mathcal{L} \setminus \{l_1\}} \operatorname{Pr}[I(\x,  M(\mathcal{D'})) = l_i]$ as:
\begin{equation}
    \operatorname{Pr}[I(\x,  M(\mathcal{D'})) = l_1] \geq \operatorname{K}_{lower}^{-1} (\operatorname{Pr}\left[I(\x,  M(\mathcal{D}))\right])
\end{equation}
and
\begin{equation}
\begin{aligned}
    \max_{l_i \in \mathcal{L} \setminus \{l_1\}} \operatorname{Pr}[I(\x,  M(\mathcal{D'})) = l_i] \leq 
    \max_{l_j \in \mathcal{L} \setminus \{l_1\}} \operatorname{K}_{upper}( \operatorname{Pr}\left[I(\x,  M(\mathcal{D})))=l_j\right])\enspace. 
\end{aligned}
\end{equation}
If the probability lower bound of predicting $l_i$ is larger than the probability upper bound of predicting any other labels
\begin{equation}
    \begin{aligned}
    \operatorname{K}_{lower}^{-1} (\operatorname{Pr}\left[I(\x,  M(\mathcal{D}))=l_1\right]) > 
     \max_{l_i \in \mathcal{L} \setminus \{l_1\}} \operatorname{K}_{upper}( \operatorname{Pr}\left[I(\x,  M(\mathcal{D}))=l_i\right])\enspace,
    \end{aligned}
\end{equation}
then we have our goal proven in the case of
\begin{equation}
    \begin{aligned}
    &\forall \mathcal{D'} \in \mathcal{B}(\mathcal{D},1), \;\;
    \operatorname{Pr}[I(\x,  M(\mathcal{D'})) = l_1] > \max_{l_i \in \mathcal{L} \setminus \{l_1\}} \operatorname{Pr}[I(\x,  M(\mathcal{D'})) = l_i] \\
    & \rightarrow l_1 = \arg\max_{l\in\mathcal{L}} \operatorname{Pr}\left[I(\x, M(\mathcal{D}))=l\right]
    \enspace. 
\end{aligned}
\end{equation}
\end{proof}

\subsection{Proof of \cref{the:Improved rdp group privacy in SGM}}\label{apd:group privacy sgm} 

\citet{mironov_renyi_dp_2019} proposed calculating the amount of Rényi-DP obtained from SGM. We extend Theorem $4$ from this ``adjacent datasets'' to ``datasets that differ in up to $r$ examples'', such that it enables the modified method to calculate group privacy of size $r$.
\begin{theorem}[Theorem 4,~\citealp{mironov_renyi_dp_2019}, for group size $r$]
\label{the:theorem 4 with group size of r}
Let $\mathrm{SG}_{q, \sigma}$ be the Sampled Gaussian mechanism for some function $f$. Then $\mathrm{SG}_{q, \sigma}$ satisfies $(\alpha, \varepsilon)$-RDP of group size $r$ whenever
\begin{equation}
\begin{aligned}
\varepsilon & \leq \mathrm{D}_{\alpha}\left(\mathcal{N}\left(0, \sigma^{2}\right) \|(1-q') \mathcal{N}\left(0, \sigma^{2}\right)+q' \mathcal{N}\left(1, \sigma^{2}\right)\right), \\
\text { and } \varepsilon & \leq \mathrm{D}_{\alpha}\left((1-q') \mathcal{N}\left(0, \sigma^{2}\right)+q' \mathcal{N}\left(1, \sigma^{2}\right) \| \mathcal{N}\left(0, \sigma^{2}\right)\right)\enspace,
\end{aligned}
\end{equation}
where $q' = 1-(1-q)^{r}$ under the assumption $\left\|f(S)-f\left(S^{\prime}\right)\right\|_{2} \leq 1$ for any $S, S^{\prime} \in \mathcal{S}$ that differ up $r$ examples.
\end{theorem}
\begin{proof}
Let $S, S^{\prime} \in \mathcal{S}$ be a pair of datasets that differ in $r$ examples, such that $S^{\prime}=S \cup\{x_1, x_2,..., x_r\}$. We wish to bound the Rényi divergences $\mathrm{D}_{\alpha}\left(\mathcal{M}(S) \|_{D} \mathcal{M}\left(S^{\prime}\right)\right)$ and $\mathrm{D}_{\alpha}\left(\mathcal{M}\left(S^{\prime}\right) \| \mathcal{M}(S)\right)$, where $\mathcal{M}$ is the Sampled Gaussian mechanism for some function $f$ with $\ell_{2}$-sensitivity $1$.

To achieve this, let $T$ denote a set-valued random variable defined by taking a random subset of $\mathcal{S}$, where each element of $S$ is independently placed in $T$ with probability $q$. Conditioned on $T$, the mechanism $\mathcal{M}(S)$ samples from a Gaussian with mean $f(T)$. Thus
\begin{equation}
    \mathcal{M}(S)=\sum_{T} p_{T} \mathcal{N}\left(f(T), \sigma^{2} \mathbb{I}^{d}\right)\enspace,
\end{equation}
where the sum here denotes mixing of the distributions with weights $p_{T}$. Similarly,
\begin{equation}
\begin{aligned}
&\mathcal{M}\left(S^{\prime}\right)=\sum_{T} p_{T}\left(\sum_{k=0}^{r}{\binom{r}{k}}
    q^{k}(1-q)^{r-k}\mathcal{N}\left(f(T \cup V), \sigma^{2} \mathbb{I}^{d}\right)\right)\\
    &V \subseteq \{x_1, x_2,..., x_r\}\\
    &k = \|V\| \enspace.
\end{aligned}
\end{equation}

As Rényi divergence is quasi-convex it is possible to construct the bound
\begin{equation}
\begin{aligned}
\mathrm{D}_{\alpha}\left(\mathcal{M}(S) \| \mathcal{M}\left(S^{\prime}\right)\right) & \leq \sup _{T} \mathrm{D}_{\alpha}\left(\mathcal{N}\left(f(T), \sigma^{2} \mathbb{I}^{d}\right) \|\sum_{k=0}^{r}{\binom{r}{k}}
    q^{k}(1-q)^{r-k}\mathcal{N}\left(f(T \cup V), \sigma^{2} \mathbb{I}^{d}\right)\right) \\
& \leq \sup _{T} \mathrm{D}_{\alpha}\left(\mathcal{N}\left(0, \sigma^{2} \mathbb{I}^{d}\right) \|\sum_{k=0}^{r}{\binom{r}{k}}
    q^{k}(1-q)^{r-k}\mathcal{N}\left(f(T \cup V) - f(T), \sigma^{2} \mathbb{I}^{d}\right)\right)\enspace,
\end{aligned}
\end{equation}
by way of the translation invariance of Rényi divergence. Since these covariances are symmetric, we can through rotation assume that $f(T \cup V) - f(T)=c_{T} \mathbf{e}_{1}$ for some constant $c_{T} \leq 1$. The two distributions at hand are then both product distributions that are identical in all coordinates except the first. By the additivity of Rényi divergence for product distributions, it then follows that
\begin{equation}
\begin{aligned}
\mathrm{D}_{\alpha}\left(\mathcal{M}(S) \| \mathcal{M}\left(S^{\prime}\right)\right) & \leq \sup _{c \leq 1} \mathrm{D}_{\alpha}\left(\mathcal{N}\left(0, \sigma^{2}\right) \|(1-q)^{r} \mathcal{N}\left(0, \sigma^{2}\right)+(1-(1-q)^{r}) \mathcal{N}\left(c, \sigma^{2}\right)\right) \\
&=\sup _{c \leq 1} \mathrm{D}_{\alpha}\left(\mathcal{N}\left(0,(\sigma / c)^{2}\right) \|(1-q)^{r} \mathcal{N}\left(0,(\sigma / c)^{2}\right)+(1-(1-q)^{r}) \mathcal{N}\left(1,(\sigma / c)^{2}\right)\right)\enspace.
\end{aligned}
\end{equation}
For any $c \leq 1$, the noise $\mathcal{N}\left(0,(\sigma / c)^{2}\right)$ can be obtained from $\mathcal{N}\left(0, \sigma^{2}\right)$ by adding noise from $\mathcal{N}\left(0,(\sigma / c)^{2}-\right.$ $\left.\sigma^{2}\right)$, and the same operation allows us to to obtain $(1-q)^{r} \mathcal{N}\left(0,(\sigma / c)^{2}\right)+ (1-(1-q)^{r}) \mathcal{N}\left(1,(\sigma / c)^{2}\right)$ from $(1-q)^{r} \mathcal{N}\left(0, \sigma^{2}\right)+ (1-(1-q)^{r}) \mathcal{N}\left(1, \sigma^{2}\right)$. Thus by the data processing inequality for Rényi divergence, we conclude
\begin{equation}
    \mathrm{D}_{\alpha}\left(\mathcal{M}(S) \| \mathcal{M}\left(S^{\prime}\right)\right) \leq \mathrm{D}_{\alpha}\left(\mathcal{N}\left(0, \sigma^{2}\right) \|(1-q)^{r} \mathcal{N}\left(0, \sigma^{2}\right)+ (1-(1-q)^{r}) \mathcal{N}\left(1, \sigma^{2}\right)\right)\enspace.
\end{equation}
An identical argument implies that
\begin{equation}
    \mathrm{D}_{\alpha}\left(\mathcal{M}\left(S^{\prime}\right) \| \mathcal{M}(S)\right) \leq \mathrm{D}_{\alpha}\left((1-q)^{r} \mathcal{N}\left(0, \sigma^{2}\right)+ (1-(1-q)^{r}) \mathcal{N}\left(1, \sigma^{2}\right) \| \mathcal{N}\left(0, \sigma^{2}\right)\right)
\end{equation}
as claimed.
\end{proof}

Note that the only difference between the original Theorem $4$ of and \cref{the:theorem 4 with group size of r} is the modified $q$ in the conclusion. To complete the proof of \cref{the:Improved rdp group privacy in SGM}, we can utilize the conclusion in \cref{the:theorem 4 with group size of r} and continue the steps after Theorem $4$ in the paper~\cite{mironov_renyi_dp_2019} by replacing $q$ with $1-(1-q)^{r}$. 

\subsection{Proof of Expected Value Bound of $(\alpha, \epsilon)$-Rényi-DP}
\label{apd:expected value rdp}
\begin{lemma}[Expected Value Bound of $(\alpha, \epsilon)$-Rényi-DP]
\label{lem:Expected Value Bound of RDP}
Suppose a randomized function $M_{DP}$, with bounded output $M_{DP}(\mathcal{D}) \in [0, b], b \in \mathbb{R}^{+}$, satisfies $(\alpha, \epsilon)$-Rényi-DP. Then for any $\mathcal{D}' \in \mathcal{B}(\mathcal{D}, 1)$ the expected value of its output follows:
\begin{equation}
\mathbb{E}(M_{DP}(\mathcal{D}')) \leq  b^{1/\alpha} (e^{\epsilon} \mathbb{E}(M_{DP}(\mathcal{D})))^{(\alpha-1)/\alpha}\enspace,
\end{equation}
where the expectation is taken over the randomness in $M_{DP}$.
\end{lemma}
\begin{proof}
We first recall Hölder's Inequality, which states that for real-valued functions $f$ and $g$, and real $p, q>1$, such that $1 / p+1 / q=1$,
\begin{equation}
\|f g\|_{1} \leq\|f\|_{p}\|g\|_{q}\enspace.
\end{equation}
This in turn allows the expected output to be expressed as 
\begin{equation}
    \mathbb{E}(M_{DP}(\mathcal{D}')) = \int_{0}^{b} \operatorname{Pr}[\mathcal{M}_{DP}(\mathcal{D}') > t] dt \enspace.
\end{equation}
By then applying the outcomes guarantee of Rényi-DP as stated in \eqref{equ:outcomes guarantee rdp}, we have that
\begin{equation}
    \mathbb{E}(M_{DP}(\mathcal{D}')) \leq \int_{0}^{b} \left(e^{\epsilon} \operatorname{Pr}\left[\mathcal{M}_{DP}\left(\mathcal{D}\right) > t \right]\right)^{(\alpha-1) / \alpha} dt \enspace.
\end{equation}
By Hölder's Inequality setting $p = \alpha$ and $q = \alpha/(\alpha - 1)$, $f(t) = 1$, $g(t) = \operatorname{Pr}\left[\mathcal{M}_{DP}\left(\mathcal{D}\right) > t \right]^{(\alpha-1) / \alpha}$, allows for us to state that
\begin{equation}
\begin{aligned}
\mathbb{E}(M_{DP}(\mathcal{D}')) & \leq e^{\epsilon (\alpha-1)/\alpha} (\int_{0}^{b} 1^{\alpha} dt)^{1/\alpha} (\int_{0}^{b} \operatorname{Pr}[\mathcal{M}_{DP}(\mathcal{D}) > t] dt)^{(\alpha-1)/\alpha} \\
    & = e^{\epsilon (\alpha-1)/\alpha} b^{1/\alpha} (\mathbb{E}(M_{DP}(\mathcal{D})))^{(\alpha-1)/\alpha} \\
    & = b^{1/\alpha} (e^{\epsilon} \mathbb{E}(M_{DP}(\mathcal{D})))^{(\alpha-1)/\alpha}\enspace.
\end{aligned}
\end{equation}
\end{proof}
Following the discussion in \cref{sec:Differential Privacy Certified Defence Framework}, if we interpret the output of $M_{DP}(\mathcal{D}', \x)$ as the returned probability distribution for each label $(y_1(\mathcal{D}', \x),..., y_{L}(\mathcal{D}', \x))$, then by applying \cref{lem:Expected Value Bound of RDP} for each label with $b=1$ it follows that we must have the expected value bound
\begin{equation}
    \mathbb{E}[y_i(\mathcal{D}',\x)] \leq (e^{\epsilon} \mathbb{E}[y_i(\mathcal{D},\x)])^{\frac{\alpha-1}{\alpha}}\enspace.
\end{equation}

\subsection{Differences in empirical results of Baseline-Bagging and Baseline-DPA}
\label{apd:mnist results diff}
We ensure all methods use the same model architecture for fair comparisons. In our experiment setting, the LeNet-5 is employed on MNIST/Fashion-MNIST, while the Opa-tut and ResNet-18 are employed on CIFAR-10. Specifically, Baseline-Bagging's original paper and code employed a simple CNN model for each base classifier on MNIST, and Baseline-DPA use the NiN~\cite{lin2013network} architecture for both MNIST and CIFAR-10. Motivated by the need to deploy such models in larger scale, practical environments, we instead considered the algorithm in the context of the the widely accepted architecture LeNet-5~\cite{lecun_gradient_based_1998} and ResNet-18~\cite{he_deep_residual_2015} for MNIST and CIFAR-10 respectively, and it was these numbers that we reported. It is these changes which have resulted in the change in relative performance.

\subsection{Improved Group Privacy in SGM Experiments}
\begin{figure}[h]
  \begin{subfigure}[b]{0.5\linewidth}
    \centering
    \includegraphics[width=\figueWidFrac\linewidth]{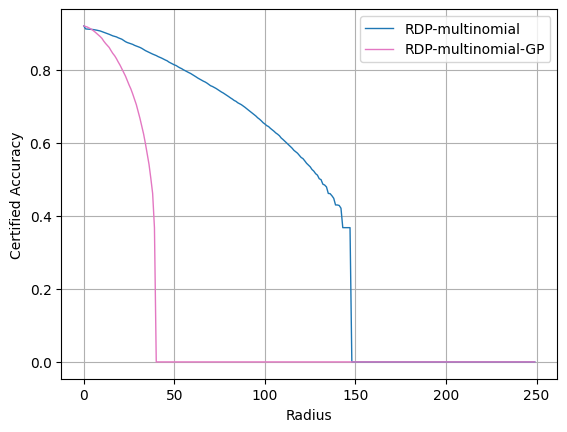} 
    \caption{MNIST (LeNet-5) against standard group privacy} 
  \end{subfigure}
  \begin{subfigure}[b]{0.5\linewidth}
    \centering
    \includegraphics[width=\figueWidFrac\linewidth]{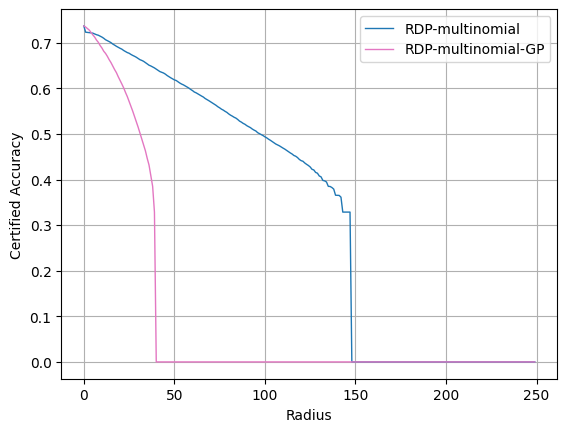} 
    \caption{Fashion-MNIST (LeNet-5), against standard group privacy}
  \end{subfigure} 
  \caption{The plots contain certified accuracy plot for the method RDP-multinomial with proposed improved group privacy (RDP-multinomial) against RDP-multinomial with standard group privacy (RDP-multinomial-GP) on datasets MNIST and Fashion-MNIST.}
  \label{fig:GP agasint Improved GP}
\end{figure}

\subsection{Certification Performance}
\label{tab:median max table}
\begin{table}[ht]
\centering
\begin{tabular}{l l c c c } 
\toprule
Dataset & Architecture                & RDP-multinomial & RDP-prob scores  & ADP-multinomial  \\ 
\cmidrule(lr){1-1} \cmidrule(lr){2-2}  \cmidrule(lr){3-5} 
MNIST                     & LeNet-5   & $127, 147$        & $94, 193$          & $35, 44$           \\ 
Fashion-MNIST             & LeNet-5   & $89, 147$         & $54, 184$          & $23, 44$           \\ 
\multirow{2}{*}{CIFAR-10} & Opa-tut   & $0, 10$           & $0, 12$            & $0, 2 $            \\ 
                          & ResNet-18 & $3, 13$           & $3, 17$            & $0, 2$             \\ 
\midrule
\multicolumn{2}{c }{}                & ADP-prob scores & Baseline-Bagging & Baseline-DPA     \\ 
 \cmidrule(lr){3-5} 
MNIST                     & LeNet-5   & $26, 70$          & $80, 80$           & $118, 125$         \\ 
Fashion-MNIST             & LeNet-5   & $14, 64$          & $57, 80$           &                  \\ 
\multirow{2}{*}{CIFAR-10} & Opa-tut   & $0, 3$            & $0, 4$             &                  \\ 
                          & ResNet-18 & $0, 8$            & $4, 8$             & $4, 7$             \\
\bottomrule
\end{tabular}
\caption{The table summarizes the \emph{median} and \emph{maximum} value of certification over all test samples for each method in each dataset. The first and second numbers in each entry represent the median and maximum value respectively.}
\end{table}

\subsection{Robustness certificate analytical form}
Consider a training dataset $\mathcal{D}$, and input instance $x$, and a randomised learner $M$. 
\paragraph{ADP-multinomial} 
Let $l_1 = \arg\max_{l\in\mathcal{L}} \operatorname{Pr}\left[I(\x, M(\mathcal{D}))=l\right]$. The pointwise-certified radius $r$ is:
\begin{align}
    &\max r\\
    &s.t.\\
    &M \text{ preserves a } (\mathcal{K}_{\epsilon, \delta}, r) \text{ outcomes guarantee }\\
    &K_{upper} = \operatorname{exp}(\epsilon_{u}) x + \delta_{u}\\
    &K_{lower} = \operatorname{exp}(\epsilon_{l}) x + \delta_{l}\\
    &\operatorname{exp}(-\epsilon_{l})(\operatorname{Pr}\left[I(\x, M(\mathcal{D}))=l_1\right] - \delta_{l}) > \max_{l \in \mathcal{L} \setminus \{l_1\}} \operatorname{exp}(\epsilon_{u})\operatorname{Pr}\left[I(\x, M(\mathcal{D}))=l\right] + \delta_{u}\\
\end{align}
\paragraph{RDP-multinomial} 
Let $l_1 = \arg\max_{l\in\mathcal{L}} \operatorname{Pr}\left[I(\x, M(\mathcal{D}))=l\right]$. The pointwise-certified radius $r$ is:
\begin{align}
    &\max r\\
    &s.t.\\
    &M \text{ preserves a } (\mathcal{K}_{\epsilon, \alpha}, r) \text{ outcomes guarantee }\\
    &K_{upper} = (\operatorname{exp}(\epsilon_{u}) x)^{\frac{\alpha_{u}-1}{\alpha_{u}}}\\
    &K_{lower} = (\operatorname{exp}(\epsilon_{l}) x)^{\frac{\alpha_{l}-1}{\alpha_{l}}}\\
    &\operatorname{exp}(-\epsilon_{l}) \operatorname{Pr}\left[I(\x, M(\mathcal{D}))=l_1\right]^{\frac{\alpha_{l}}{\alpha_{l}-1}} > \max_{l \in \mathcal{L} \setminus \{l_1\}}
    (\operatorname{exp}(\epsilon_{u}) \operatorname{Pr}\left[I(\x, M(\mathcal{D}))=l\right])^{\frac{\alpha_{u}-1}{\alpha_{u}}}
\end{align}
\paragraph{ADP-probability scores} 
Let $l_1 = \arg\max_{l\in\mathcal{L}} \mathbb{E}[ y_l(M(\mathcal{D}), \x)]$. The pointwise-certified radius $r$ is:
\begin{align}
    &\max r\\
    &s.t.\\
    &M \text{ preserves a } (\mathcal{K}_{\epsilon, \delta}, r) \text{ outcomes guarantee }\\
    &K_{upper} = \operatorname{exp}(\epsilon_{u}) x + \delta_{u}\\
    &K_{lower} = \operatorname{exp}(\epsilon_{l}) x + \delta_{l}\\
    &\operatorname{exp}(-\epsilon_{l})(\mathbb{E}[ y_{l_{1}}(M(\mathcal{D}), \x)] - \delta_{l}) > \max_{l \in \mathcal{L} \setminus \{l_1\}} \operatorname{exp}(\epsilon_{u})\mathbb{E}[ y_l(M(\mathcal{D}), \x)] + \delta_{u}\\
\end{align}
\paragraph{RDP-probability scores} 
Let $l_1 = \arg\max_{l\in\mathcal{L}} \mathbb{E}[ y_l(M(\mathcal{D}), \x)]$. The pointwise-certified radius $r$ is:
\begin{align}
    &\max r\\
    &s.t.\\
    &M \text{ preserves a } (\mathcal{K}_{\epsilon, \alpha}, r) \text{ outcomes guarantee }\\
    &K_{upper} = (\operatorname{exp}(\epsilon_{u}) x)^{\frac{\alpha_{u}-1}{\alpha_{u}}}\\
    &K_{lower} = (\operatorname{exp}(\epsilon_{l}) x)^{\frac{\alpha_{l}-1}{\alpha_{l}}}\\
    &\operatorname{exp}(-\epsilon_{l}) \mathbb{E}[ y_{l_{1}}(M(\mathcal{D}), \x)]^{\frac{\alpha_{l}}{\alpha_{l}-1}} > \max_{l \in \mathcal{L} \setminus \{l_1\}}
    (\operatorname{exp}(\epsilon_{u}) \mathbb{E}[ y_l(M(\mathcal{D}), \x)])^{\frac{\alpha_{u}-1}{\alpha_{u}}}
\end{align}



\end{document}